\makeatletter
\@namedef{ver@everyshi.sty}{}
\makeatother

\documentclass[final]{cvpr}

\usepackage{times}
\usepackage{epsfig}
\usepackage{graphicx}
\usepackage{amsmath}
\usepackage{amssymb}
\usepackage{svg}
\usepackage{gensymb}
\usepackage{siunitx}
\usepackage{multirow}
\usepackage{hhline}
\usepackage[super]{nth}
\usepackage{amssymb}
\usepackage{url}
\usepackage{xcolor,colortbl}
\usepackage{array}
\usepackage{booktabs}
\usepackage{stfloats}
\newcolumntype{x}[1]{>{\centering\arraybackslash\hspace{0pt}}p{#1}}

\newcommand{\method}{CaDDN}

\usepackage[pagebackref=true,breaklinks=true,colorlinks,bookmarks=false]{hyperref}
\makeatletter
\newcommand\newtag[2]{#1\def\@currentlabel{#1}\label{#2}}
\makeatother
\definecolor{TableBlue}{HTML}{E0FFFF}

\def\cvprPaperID{3916} 
\def\confYear{CVPR 2021}

\begin{document}
\title{Categorical Depth Distribution Network for Monocular 3D Object Detection}
\author{
Cody Reading \ \ \ \ Ali Harakeh \ \ \ \ Julia Chae \ \ \ \ Steven L. Waslander\\
University of Toronto Robotics Institute\\
{\tt\small \{cody.reading, nayoung.chae\}@mail.utoronto.ca, ali.harakeh@utoronto.ca, stevenw@utias.utoronto.ca}
}
\maketitle
\begin{abstract}
Monocular 3D object detection is a key problem for autonomous vehicles, as it provides a solution with simple configuration compared to typical multi-sensor systems. The main challenge in monocular 3D detection lies in accurately predicting object depth, which must be inferred from object and scene cues due to the lack of direct range measurement. Many methods attempt to directly estimate depth to assist in 3D detection, but show limited performance as a result of depth inaccuracy. Our proposed solution, Categorical Depth Distribution Network (\method), uses a predicted categorical depth distribution for each pixel to project rich contextual feature information to the appropriate depth interval in 3D space. We then use the computationally efficient bird's-eye-view projection and single-stage detector to produce the final output detections. We design \method~as a fully differentiable end-to-end approach for joint depth estimation and object detection. We validate our approach on the KITTI 3D object detection benchmark,  where we rank \nth{1} among published monocular methods. We also provide the first monocular 3D detection results on the newly released Waymo Open Dataset. We provide a code release for \method~which is made available \href{https://github.com/TRAILab/CaDDN}{here}.
\end{abstract}
\begin{figure}
\begin{center}
\includesvg[width=\columnwidth]{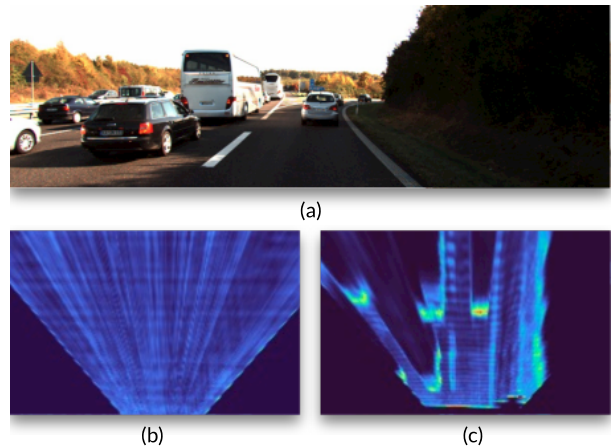}
\end{center}
\vspace{-3mm}
 \caption{(a) Input image. (b) Without depth distribution supervision, BEV features from \method~suffer from smearing effects. (c) Depth distribution supervision encourages BEV features from \method~ to encode meaningful depth confidence, in which objects can be accurately detected.}
\label{fig:intro}
\end{figure}
\vspace{-4mm}
\section{Introduction} \label{sec:Introduction}
Perception in 3D space is a key component in fields such as autonomous vehicles and robotics, enabling systems to understand their environment and react accordingly. LiDAR~\cite{AVOD, PV-RCNN, PointRCNN} and stereo~\cite{e2e_pseudo-lidar, OCStereo, CGStereo, DSGN} sensors have a long history of use for 3D perception tasks, showing excellent results on 3D object detection benchmarks such as the KITTI 3D object detection benchmark~\cite{Kitti} due to their ability to generate precise 3D measurements.

Monocular based 3D perception has been pursued simultaneously, motivated by the potential for a low-cost, easy-to-deploy solution with a single camera~\cite{Mono3D, Deep3DBox, DeepMANTA, monopsr}.  Performance on the same 3D object detection benchmarks lags significantly relative to LiDAR and stereo methods, due to the loss of depth information when scene information is projected onto the image plane.

To combat this effect, monocular object detection methods~\cite{D4LCN, PatchNet, AM3D, Psuedo-LIDAR} often learn depth explicitly, by training a monocular depth estimation network in a separate stage. However, depth estimates are consumed directly in the 3D object detection stage without an understanding of depth confidence, leading to networks that tend to be overconfident in depth predictions. Over-confidence in depth is particularly an issue at long range~\cite{Psuedo-LIDAR}, leading to poor localization. Further, depth estimation is separated from 3D detection during the training phase, preventing depth map estimates from being optimized for the detection task.

Depth information in image data can also be learned implicitly, by directly transforming features from images to 3D space and finally to bird's-eye-view (BEV) grids~\cite{OFT, LiftSplatShoot}. Implicit methods, however, tend to suffer from feature smearing, wherein similar image features can exist at multiple locations in the projected space. Feature smearing increases the difficulty of localizing objects in the scene.

To resolve the identified issues, we propose a monocular 3D object detection method, \method, that enables accurate 3D detection by learning categorical depth distributions. By leveraging probabilistic depth estimation, \method~is able to generate high quality bird's-eye-view feature representations from images in an end-to-end fashion. We summarize our approach with three contributions.
\\[0.2\baselineskip]
\noindent\textbf{(1) Categorical Depth Distributions.} In order to perform 3D detection, we predict pixel-wise categorical depth distributions to accurately locate image information in 3D space. Each predicted distribution describes the probabilities that a pixel belongs to a set of predefined depth bins. We encourage our distributions to be as sharp as possible around the correct depth bins, in order to encourage our network to focus more on image information where depth estimation is both accurate and confident~\cite{Sharpness_Proof}. By doing so, our network is able to produce sharper and more accurate features that are useful for 3D detection (see Figure~\ref{fig:intro}). On the other hand, our network retains the ability to produce less sharp distributions when depth estimation confidence is low. Using categorical distributions allows our feature encoding to capture the inherent depth estimation uncertainty to reduce the impact of erroneous depth estimates, a property shown to be key to \method's improved performance in Section~\ref{sec:Ablation Studies}. Sharpness in our predicted depth distributions is encouraged through supervision with one-hot encodings of the correct depth bin, which can be generated by projecting LiDAR depth data into the camera frame.
\\[0.2\baselineskip]
\noindent\textbf{(2) End-To-End Depth Reasoning.} We learn depth distributions in an end-to-end fashion, jointly optimizing for accurate depth prediction as well as accurate 3D object detection. We argue that joint depth estimation and 3D detection reasoning encourages depth estimates to be optimized for the 3D detection task, leading to increased performance as shown in Section~\ref{sec:Ablation Studies}.
\\[0.2\baselineskip]
\noindent\textbf{(3) BEV Scene Representation.} We introduce a novel method to generate high quality bird's-eye-view scene representations from single images using categorical depth distributions and projective geometry. We select the bird's-eye-view representation due to its ability to produce excellent 3D detection performance with high computational efficiency~\cite{PointPillars}. The generated bird's-eye-view representation is used as input to a bird's-eye-view based detector to produce the final output.
\\[0.2\baselineskip]
\indent
\method~is shown to rank first among all previously published monocular methods on the Car and Pedestrian categories of the KITTI 3D object detection test benchmark~\cite{KITTI_Test}, with margins of $1.69\%$ and $1.46\%$ $\mathrm{AP}|_{R_{40}}$ respectively. We are the first to report monocular 3D object detection results on the Waymo Open Dataset~\cite{waymo}.
\section{Related Work} \label{sec:RelatedWork}
\noindent\textbf{Monocular Depth Estimation}.
Monocular depth estimation is performed by generating a single depth value for every pixel in an image. As such, many monocular depth estimation methods are based on architectures used in well-studied pixel-to-pixel mapping problems such as semantic segmentation. As an example, fully convolutional networks (FCNs)~\cite{FCN} were introduced for semantic segmentation, and were subsequently adopted for monocular depth estimation~\cite{Laina}. The atrous spatial pyramid pooling (ASPP) module was also first proposed for semantic segmentation in DeepLab~\cite{DeepLab, DeepLabAlt, DeepLabV3} and subsequently used for depth estimation in DORN~\cite{DORN} and BTS~\cite{BTS}. Further, many methods jointly perform depth estimation and segmentation~\cite{PAD-Net, JointSeg, SDC-Depth, Eigen2} in an end-to-end manner. We follow the design of the semantic segmentation network DeepLabV3~\cite{DeepLabV3} for estimating categorical depth distributions for each pixel in the image.
\\[0.2\baselineskip]
\noindent\textbf{BEV Semantic Segmentation}.
BEV segmentation methods~\cite{CrossView, Learning} attempt to predict BEV semantic maps of 3D scenes from images. Images can be used to either directly estimate BEV semantic maps~\cite{MonoLayout, VAEMap, TopView} or to estimate a BEV feature representation~\cite{LiftSplatShoot, PON, bevseg} as an intermediate step for the segmentation task.  In particular, Lift, Splat, Shoot~\cite{LiftSplatShoot} predicts categorical depth distributions in an unsupervised manner, in order to generate intermediate BEV representations. In this work, we predict categorical depth distributions via supervision with ground truth one-hot encodings to generate more accurate depth distributions for object detection.
\begin{figure*}
\begin{center}
\includesvg[width=\textwidth]{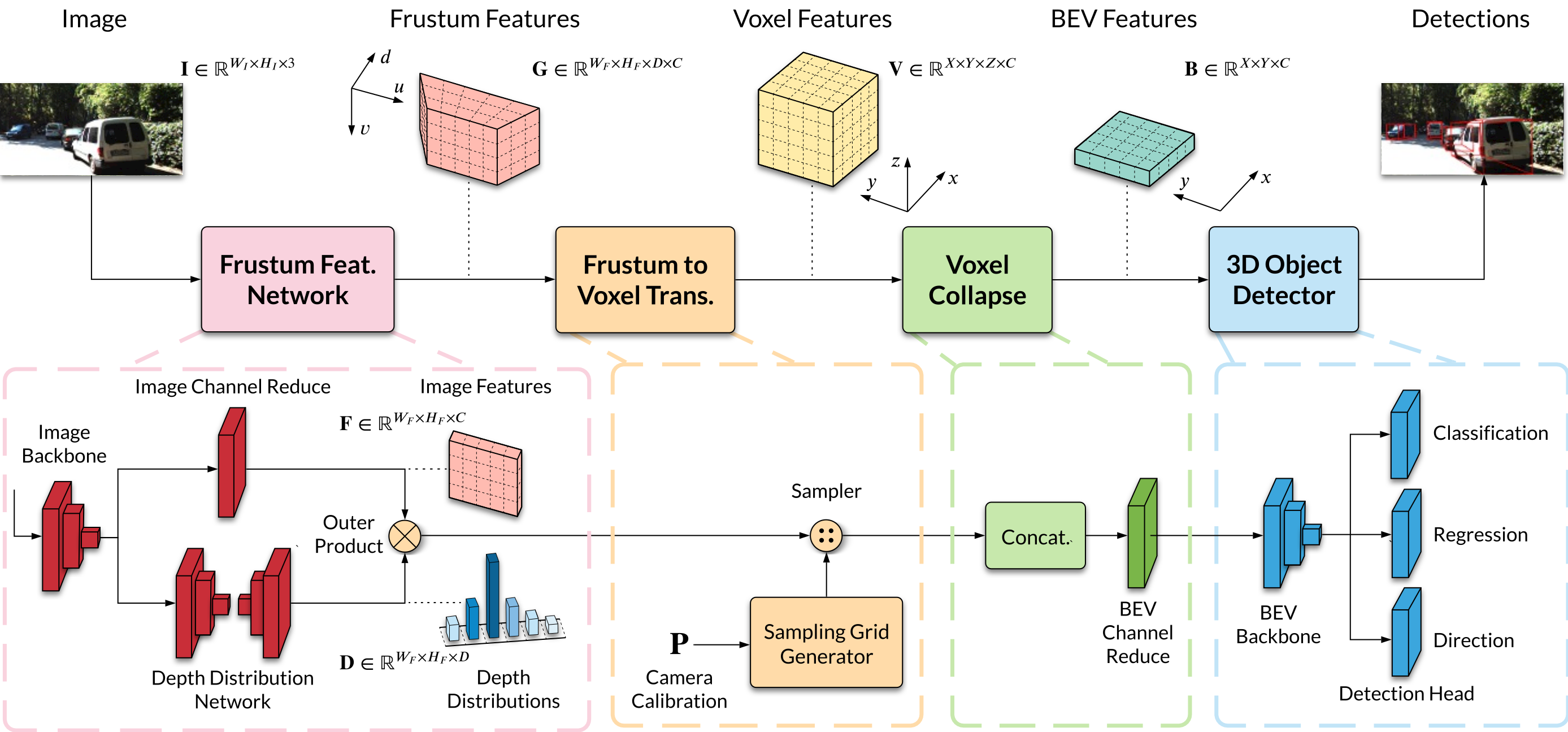}
\end{center}
   \caption{\method~Architecture. The network is composed of three modules to generate 3D feature representations and one to perform 3D detection. Frustum features $\mathbf{G}$ are generated from an image $\mathbf{I}$ using estimated depth distributions $\mathbf{D}$, which are transformed into voxel features $\mathbf{V}$. The voxel features are collapsed to bird's-eye-view features $\mathbf{B}$ to be used for 3D object detection.}
\label{fig:architecture}
\end{figure*}
\\[0.2\baselineskip]
\noindent\textbf{Monocular 3D Detection}.
Monocular 3D object detection methods often generate intermediate representations to assist in the 3D detection task. Based on these representations, monocular detection can be divided into three categories: direct, depth-based, and grid-based methods.
\\[0.2\baselineskip]
\noindent\textit{Direct Methods}.
Direct methods~\cite{Mono3D, MonoDis, Kinematic3D, RARNet} estimate 3D detections directly from images without predicting an intermediate 3D scene representation. Rather, direct methods~\cite{MoVi-3D, MonoPair, Deep3DBox, M3D-RPN} can incorporate the geometric relationship between the 2D image plane and 3D space to assist with detections. For example, object keypoints can be estimated on the image plane, in order to assist in 3D box construction using known geometry~\cite{SMOKE, RTM3D}. M3D-RPN~\cite{M3D-RPN} introduces depth-aware convolutions that divides the input row-wise and learns non-shared kernels for each region, to learn location specific features that correlate to regions in 3D space. Shape estimation can be performed for objects in the scene to create an understanding of 3D object geometry. Shape estimates can be supervised from labeled vertices of 3D CAD models~\cite{DeepMANTA, 3DRCNN}, from LiDAR scans~\cite{monopsr}, or directly from input data in a self-supervised manner~\cite{DiffRendering}. A drawback for direct methods is that detections are generated directly from 2D images, without access to explicit depth information, usually resulting in reduced performance in localization relative to other methods.
\\[0.2\baselineskip]
\noindent\textit{Depth-Based Methods}.
Depth-based methods perform the 3D detection task using pixel-wise depth maps as an additional input, where the depth maps are precomputed using monocular depth estimation architectures~\cite{DORN}. Estimated depth maps can be used in combination with images to perform the 3D detection task~\cite{ROI-10D,UR3D,PatchNet,D4LCN}. Alternatively, depth maps can be converted to 3D point clouds, commonly known as Pseudo-LiDAR~\cite{Psuedo-LIDAR}, which are either used directly ~\cite{Mono3D_Plidar, DA-3Ddet} or combined with image information~\cite{MultiFusion,AM3D} to generate 3D object detection results. Depth-based methods separate depth estimation from 3D object detection during the training stage, leading to the learning of sub-optimal depth maps used for the 3D detection task. Accurate depth should be prioritized for pixels belonging to objects of interest, and is less important for background pixels, a property that is not captured if depth estimation and object detection are trained independently.
\\[0.2\baselineskip]
\noindent\textit{Grid-Based Methods.}
Grid-based methods avoid estimating raw depth values by predicting a BEV grid~\cite{OFT, BirdGAN} representation, to be used as input for 3D detection architectures. Specifically, OFT~\cite{OFT} populates a voxel grid by projecting voxels into the image plane and sampling image features, to be transformed into a BEV representation. Multiple voxels can be projected to the same image feature, leading to repeated features along the projection ray and reduced detection accuracy.

\method~addresses all identified issues by jointly performing depth estimation and 3D object detection in an end-to-end manner, and leverages the depth estimates to generate meaningful bird's-eye-view representations with accurate and localized features.
\section{Methodology} \label{Method}
\method~learns to generate BEV representations from images by projecting image features into 3D space. 3D object detection is then performed with the rich BEV representation using an efficient BEV detection network. An overview of \method's architecture is shown in Figure~\ref{fig:architecture}.
\subsection{3D Representation Learning} \label{sec:3D Representation Learning}
Our network learns to produce BEV representations that are well-suited for the task of 3D object detection. Taking an image as input, we construct a frustum feature grid using the estimated categorical depth distributions. The frustum feature grid is transformed into a voxel grid using known camera calibration parameters, and then collapsed to a bird's-eye-view feature grid.
\begin{figure}
\begin{center}
\includesvg[width=\columnwidth]{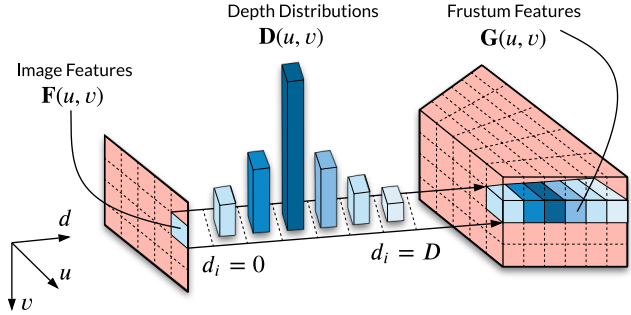}
\end{center}
 \caption{Each feature pixel $\mathbf{F}(u, v)$ is weighted by its depth distribution probabilities $\mathbf{D}(u, v)$ of belonging to $D$ discrete depth bins to generate frustum features $\mathbf{G}(u, v)$.}
\label{fig:frustum}
\end{figure}
\\[0.2\baselineskip]
\noindent\textbf{Frustum Feature Network}.
The purpose of the frustum feature network is to project image information into 3D space, by associating image features to estimated depths. Specifically, the input to the frustum feature network is an image $\mathbf{I} \in \mathbb{R}^{W_I \times H_I \times 3}$, where $W_I, H_I$ are the width and height of the image. The output is a frustum feature grid $\mathbf{G} \in \mathbb{R}^{W_F \times H_F \times D \times C}$, where $W_F, H_F,$ are the width and height of the image feature representation, $D$ is the number of discretized depth bins, and $C$ is the number of feature channels. We note that the structure of the frustum grid is similar to the plane-sweep volume used in the stereo 3D detection method DSGN~\cite{DSGN}.

A ResNet-101~\cite{ResNet} backbone is used to extract image features $\mathbf{\tilde{F}} \in \mathbb{R}^{W_F \times H_F \times C}$ (see Image Backbone in Figure~\ref{fig:architecture}). In our implementation, we extract the image features from \textit{Block1} of the ResNet-101 backbone in order to maintain a high spatial resolution. A high spatial resolution is necessary for an effective frustum to voxel grid transformation, such that the frustum grid can be finely sampled without repeated features.

The image features $\mathbf{\tilde{F}}$ are used to estimate pixel-wise categorical depth distributions $\mathbf{D} \in \mathbb{R}^{W_F \times H_F \times D}$, where the categories are the $D$ discretized depth bins. Specifically, we predict $D$ probabilities for each pixel in the image features $\mathbf{\tilde{F}}$, where each probability indicates the network's confidence that depth value belongs to a specified depth bin. The definition of the depth bins relies on the depth discretization method as discussed in Section~\ref{sec:Depth Discretization}.

We follow the design of the semantic segmentation network DeepLabV3~\cite{DeepLabV3} to estimate the categorical depth distributions from image features $\mathbf{\tilde{F}}$ (Depth Distribution Network in Figure~\ref{fig:architecture}), where we modify the network to produce pixel-wise probability scores of belonging to depth bins rather than semantic classes with a downsample-upsample architecture. Image features $\mathbf{\tilde{F}}$ are downsampled with the remaining components of the ResNet-101~\cite{ResNet} backbone (\textit{Block2}, \textit{Block3}, and \textit{Block4}). An atrous spatial pyramid pooling~\cite{DeepLabV3} (ASPP) module is applied to capture multi-scale information, where the number of output channels is set as $D$. The output of the ASPP module is upsampled to the original feature size with bilinear interpolation to produce the categorical depth distributions $\mathbf{D} \in \mathbb{R}^{W_F \times H_F \times D}$. A softmax function is applied for each pixel to normalize the $D$ logits into probabilities between $0$ and $1$.

In parallel to estimating depth distributions, we perform channel reduction (Image Channel Reduce in Figure~\ref{fig:architecture}) on image features $\mathbf{\tilde{F}}$  to generate the final image features $\mathbf{F}$, using a 1x1 convolution + BatchNorm + ReLU layer to reduce the number of channels from $C = 256$ to $C = 64$. Channel reduction is required to reduce the high memory footprint of ResNet-101 features that will be populated in the 3D frustum grid.

Let $(u, v, c)$ represent a coordinate in image features $\mathbf{F}$ and $(u, v, d_i)$ represent a coordinate in categorical depth distributions $\mathbf{D}$, where $(u, v)$ are the feature pixel location, $c$ is the channel index, and $d_i$ is the depth bin index. To generate a frustum feature grid $\mathbf{G}$, each feature pixel $\mathbf{F}(u, v)$ is weighted by its associated depth bin probabilities in $\mathbf{D}(u, v)$ to populate the depth axis $d_i$, visualized in Figure~\ref{fig:frustum}. Feature pixels can be weighted by depth probability using the outer product, defined as:
\begin{align}
    \label{eq:outer_prod}
    \mathbf{G}(u, v) &= \mathbf{D}(u, v) \otimes \mathbf{F}(u, v)
\end{align}
\noindent
where $\mathbf{D}(u, v)$ is the predicted depth distribution and $\mathbf{G}(u, v)$ is an output matrix of size $D \times C$. The outer product in Equation~\ref{eq:outer_prod} is computed for each pixel to form frustum features $\mathbf{G} \in \mathbb{R}^{W_F \times H_F \times D \times C}$.
\begin{figure}
\begin{center}
\includesvg[width=\columnwidth]{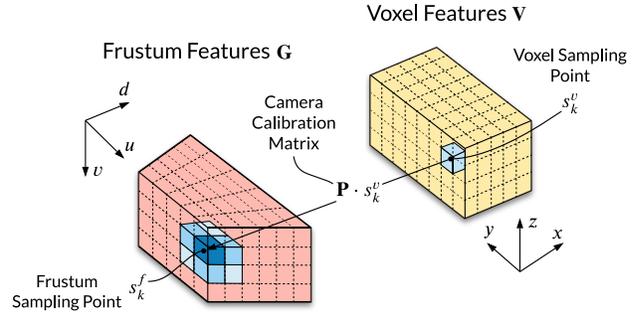}
\end{center}
 \caption{Sampling points in each voxel are projected into the frustum grid. Frustum features are sampled using trilinear interpolation (shown as blue in $\mathbf{G}$) to populate voxels in $\mathbf{V}$.}
\label{fig:frustum2voxel}
\end{figure}
\\[0.2\baselineskip]
\noindent\textbf{Frustum to Voxel Transformation}.
The frustum features $\mathbf{G} \in \mathbb{R}^{W_F \times H_F \times D \times C}$ are transformed to a voxel representation $\mathbf{V} \in \mathbb{R}^{X \times Y \times Z \times C}$ leveraging known camera calibration and differentiable sampling, shown in Figure~\ref{fig:frustum2voxel}. Voxel sampling points $s_k^v = [x, y, z]_k^T$ are generated at the center of each voxel and transformed to the frustum grid to form frustum sampling points $\tilde{s}_k^f = [u, v, d_c]_k^T$, where $d_c$ is the continuous depth value along the frustum depth axis $d_i$. The transformation is performed using the camera calibration matrix $\mathbf{P} \in \mathbb{R}^{3 \times 4}$. Each continuous depth value $d_c$ is converted to a discrete depth bin index $d_i$ using the depth discretization method outlined in Section~\ref{sec:Depth Discretization}. Frustum features in $\mathbf{G}$ are sampled using sampling points $s_k^f = [u, v, d_i]_k^T$ with trilinear interpolation (shown in blue in Figure~\ref{fig:frustum2voxel}) to populate voxel features in $\mathbf{V}$.

The spatial resolution of the frustum grid $\mathbf{G}$ and the voxel grid $\mathbf{V}$ should be similar for an effective transformation.
A high resolution voxel grid $\mathbf{V}$ leads to a high density of sampling points that will oversample a low resolution frustum grid, resulting in a large amount of similar voxel features. Therefore, we extract the features $\mathbf{\tilde{F}}$ from \textit{Block1} of the ResNet-101 backbone to ensure our frustum grid $\mathbf{G}$ is of high spatial resolution.
\\[0.2\baselineskip]
\noindent \textbf{Voxel Collapse to BEV}.
The voxel features $\mathbf{V} \in \mathbb{R}^{X \times Y \times Z \times C}$ are collapsed to a single height plane to generate bird's-eye-view features $\mathbf{B} \in \mathbb{R}^{X \times Y \times C}$. BEV grids greatly reduce the computational overhead while offering similar detection performance to 3D voxel grids~\cite{PointPillars}, motivating their use in our network. We concatenate the vertical axis $z$ of the voxel grid $\mathbf{V}$ along the channel dimension $c$ to form a BEV grid $\mathbf{\tilde{B}} \in \mathbb{R}^{X \times Y \times Z*C}$. The number of channels are reduced using a 1x1 convolution + BatchNorm + ReLU layer (see BEV Channel Reduce in Figure~\ref{fig:architecture}), which retrieves the original number of channels $C$ while learning the relative importance of each height slice, resulting in a BEV grid $\mathbf{B} \in \mathbb{R}^{X \times Y \times C}$.
\subsection{BEV 3D Object Detection} \label{BEV 3D Object Detection}
To perform 3D object detection on the BEV feature grid, we adopt the backbone and detection head of the well-established BEV 3D object detector PointPillars~\cite{PointPillars}, as it has been shown to provide accurate 3D detection results with a low computational overhead.  For the BEV backbone, we increase the number of 3x3 convolution + BatchNorm + ReLU layers in the downsample blocks from (4, 6, 6) used in the original PointPillars~\cite{PointPillars} to (10, 10, 10) for \textit{Block1}, \textit{Block2}, and \textit{Block3} respectively. Increasing the number of convolutional layers expands the learning capacity in our BEV network, important for learning from lower quality features produced by images compared to higher quality features originally produced by LiDAR point clouds. We use the same detection head as PointPillars~\cite{PointPillars} to generate our final detections.
\subsection{Depth Discretization} \label{sec:Depth Discretization}
\begin{figure}
\begin{center}
\includesvg[width=1.16\columnwidth]{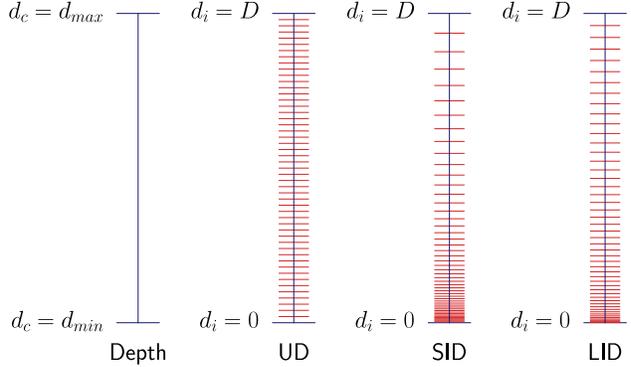}
\end{center}
 \caption{Depth Discretization Methods. Depth $d_c$ is discretized over a depth range $[d_{min}, d_{max}]$ into $D$ discrete bins. Commonly used methods include uniform (UD), spacing-increasing (SID), and linear-increasing (LID) discretization.}
\label{fig:dist}
\end{figure}

The continuous depth space is discretized in order to define the set of $D$ bins used in the depth distributions $\mathbf{D}$. Depth discretization can be performed with uniform discretization (UD) with a fixed bin size, spacing-increasing discretization (SID)~\cite{DORN}  with increasing bin sizes in $\log$ space, or linear-increasing discretization (LID)~\cite{Center3D} with linearly increasing bin sizes. Depth discretization techniques are visualized in Figure~\ref{fig:dist}.
 We adopt LID as our depth discretization as it provides balanced depth estimation for all depths~\cite{Center3D}. LID is defined as:
\begin{align} \label{eq:disc}
     d_c &= d_{\min} + \frac{d_{\max}-d_{\min}}{D(D + 1)} \cdot d_i (d_i + 1)
\end{align}
\noindent
where $d_c$ is the continuous depth value, $[d_{\min}, d_{\max}]$ is the full depth range to be discretized, $D$ is the number of depth bins, and $d_i$ is the depth bin index.
\begin{table*}
\centering
\begin{tabular}{p{3.6cm}|c|x{0.9cm}x{0.9cm}x{0.9cm}|x{0.9cm}x{0.9cm}x{0.9cm}|x{0.9cm}x{0.9cm}x{0.9cm}}
\toprule
 &  & \multicolumn{3}{c|}{Car (IOU = 0.7)} & \multicolumn{3}{c|}{Pedestrian (IOU = 0.5)} & \multicolumn{3}{c}{Cyclist (IOU = 0.5)} \\
\multirow{-2}{*}{Method} & \multirow{-2}{*}{Frames} & Easy & \textbf{Mod.} & Hard & Easy & \textbf{Mod.} & Hard & Easy & \textbf{Mod.} & Hard \\ \hline
Kinematic3D~\cite{Kinematic3D} & 4 & 19.07 & 12.72 & 9.17 & -- & -- & -- & -- & -- & -- \\ \hline
OFT~\cite{OFT} & 1 & 1.61 & 1.32 & 1.00 & 0.63 & 0.36 & 0.35 & 0.14 & 0.06 & 0.07 \\
ROI-10D~\cite{ROI-10D} & 1 & 4.32 & 2.02 & 1.46 & -- & -- & -- & -- & -- & -- \\
MonoPSR~\cite{monopsr} & 1 & 10.76 & 7.25 & 5.85 & 6.12 & 4.00 & 3.30 & {\color[HTML]{FE0000} \textbf{8.37}} & {\color[HTML]{FE0000} \textbf{4.74}} & {\color[HTML]{FE0000} \textbf{3.68}} \\
Mono3D-PLiDAR~\cite{Mono3D_Plidar} & 1 & 10.76 & 7.50 & 6.10 & -- & -- & -- & -- & -- & -- \\
MonoDIS~\cite{MonoDis} & 1 & 10.37 & 7.94 & 6.40 & -- & -- & -- & -- & -- & -- \\
UR3D~\cite{UR3D} & 1 & 15.58 & 8.61 & 6.00 & -- & -- & -- & -- & -- & -- \\
M3D-RPN~\cite{M3D-RPN} & 1 & 14.76 & 9.71 & 7.42 & 4.92 & 3.48 & 2.94 & 0.94 & 0.65 & 0.47 \\
SMOKE~\cite{SMOKE} & 1 & 14.03 & 9.76 & 7.84 & -- & -- & -- & -- & -- & -- \\
MonoPair~\cite{MonoPair} & 1 & 13.04 & 9.99 & 8.65 & {\color[HTML]{3531FF} \textbf{10.02}} & {\color[HTML]{3531FF} \textbf{6.68}} & {\color[HTML]{3531FF} \textbf{5.53}} & 3.79 & 2.12 & 1.83 \\
RTM3D~\cite{RTM3D} & 1 & 14.41 & 10.34 & 8.77 & -- & -- & -- & -- & -- & -- \\
AM3D~\cite{AM3D} & 1 & 16.50 & 10.74 & 9.52 & -- & -- & -- & -- & -- & -- \\
MoVi-3D~\cite{MoVi-3D} & 1 & 15.19 & 10.90 & 9.26 & 8.99 & 5.44 & 4.57 & 1.08 & 0.63 & 0.70 \\
RAR-Net~\cite{RARNet} & 1 & 16.37 & 11.01 & 9.52 & -- & -- & -- & -- & -- & -- \\
PatchNet~\cite{PatchNet} & 1 & 15.68 & 11.12 & {\color[HTML]{3531FF} \textbf{10.17}} & -- & -- & -- & -- & -- & -- \\
DA-3Ddet~\cite{DA-3Ddet} & 1 & {\color[HTML]{3531FF} \textbf{16.77}} & 11.50 & 8.93 & -- & -- & -- & -- & -- & -- \\
D4LCN~\cite{D4LCN} & 1 & 16.65 & {\color[HTML]{3531FF} \textbf{11.72}} & 9.51 & 4.55 & 3.42 & 2.83 & 2.45 & 1.67 & 1.36 \\ \hline
\textbf{\method~(ours)} & 1 & {\color[HTML]{FE0000} \textbf{19.17}} & {\color[HTML]{FE0000} \textbf{13.41}} & {\color[HTML]{FE0000} \textbf{11.46}} & {\color[HTML]{FE0000} \textbf{12.87}} & {\color[HTML]{FE0000} \textbf{8.14}} & {\color[HTML]{FE0000} \textbf{6.76}} & {\color[HTML]{3531FF} \textbf{7.00}} & {\color[HTML]{3531FF} \textbf{3.41}} & {\color[HTML]{3531FF} \textbf{3.30}} \\
\cellcolor{TableBlue}\textit{Improvement} & \cellcolor{TableBlue}-- & \cellcolor{TableBlue}\textit{+2.40} & \cellcolor{TableBlue}\textit{+1.69} & \cellcolor{TableBlue}\textit{+1.29} & \cellcolor{TableBlue}\textit{+2.85} & \cellcolor{TableBlue}\textit{+1.46} & \cellcolor{TableBlue}\textit{+1.23} & \cellcolor{TableBlue}\textit{-1.37} & \cellcolor{TableBlue}\textit{-1.33} & \cellcolor{TableBlue}\textit{-0.38} \\ \bottomrule
\end{tabular}
\caption{3D detection results on the KITTI~\cite{Kitti} \textit{test} set. Results are shown using the $\mathrm{AP}|_{R_{40}}$ metric only for results that are readily available. We indicate the highest result with {\color[HTML]{FE0000} \textbf{red}} and the second highest with {\color[HTML]{3531FF} \textbf{blue}}. Full results for \method~can be accessed \href{http://www.cvlibs.net/datasets/kitti/eval_object_detail.php?&result=d593e0abe5e5f8b0a0ea1523816e7851b0266cd8}{here}.}
\label{tab:kitti-test}
\end{table*}
\subsection{Depth Distribution Label Generation} \label{sec:DepthLabel}
We require depth distribution labels $\mathbf{\hat{D}}$ in order to supervise our predicted depth distributions. Depth distribution labels are generated by projecting LiDAR point clouds into the image frame to create sparse dense maps. Depth completion~\cite{IPBasic} is performed to generate depth values at each pixel in the image. We require depth information at each image feature pixel, so we downsample the depth maps of size $W_I \times H_I$ to the image feature size $W_F \times H_F$. The depth maps are converted to bin indices using the LID discretization method described in Section~\ref{sec:Depth Discretization}, followed by a conversion into a one-hot encoding to generate the depth distribution labels. A one-hot encoding ensures the depth distribution labels are sharp, essential to encourage sharpness in our depth distribution predictions via supervision.
\subsection{Training Losses} \label{sec:TrainingLosses}
Generally, classification is performed by predicting categorical distributions, and encouraging sharpness in the distribution in order to select the correct class~\cite{Class}. We leverage classification to encourage a single correct depth bin when supervising the depth distribution network, using the focal loss~\cite{FocalLoss}:
\begin{align}
    L_{\mathrm{depth}}=\frac{1}{W_F \cdot H_F}\sum_{u=1}^{W_F}\sum_{v=1}^{H_F}\mathrm{FL}(\mathbf{D}(u, v), \mathbf{\hat{D}}(u, v))
\end{align}
where $\mathbf{D}$ is the depth distribution predictions and $\mathbf{\hat{D}}$ is the depth distribution labels. We found that autonomous driving datasets contain images with fewer object pixels than background pixels, leading to loss functions that prioritize background pixels when all pixel losses are weighted evenly. We set the focal loss~\cite{FocalLoss} weighting factor $\alpha$ as  $\alpha_{\mathrm{fg}}= 3.25$ for foreground object pixels and $\alpha_{\mathrm{bg}} = 0.25$ for background pixels. Foreground object pixels are determined as all pixels that lie within 2D object bounding box labels, and background pixels are all remaining pixels. We set the focal loss~\cite{FocalLoss} focusing parameter $\gamma = 2.0$.

We use the classification loss $L_{\mathrm{cls}}$, regression loss $L_{\mathrm{reg}}$, and direction classification loss $L_{\mathrm{dir}}$ from PointPillars~\cite{PointPillars} for 3D object detection. The total loss of our network is the combination of the depth and 3D detection losses:
\begin{align}
    \label{eq:total_loss}
    L = \lambda_{\mathrm{depth}}L_{\mathrm{depth}} + \lambda_{\mathrm{cls}}L_{\mathrm{cls}} + \lambda_{\mathrm{reg}}L_{\mathrm{reg}} +
    \lambda_{\mathrm{dir}}L_{\mathrm{dir}}
\end{align}
\noindent
where $\lambda_{\mathrm{depth}}, \lambda_{\mathrm{cls}}, \lambda_{\mathrm{reg}}, \lambda_{\mathrm{dir}}$ are fixed loss weighting factors.
\section{Experimental Results} \label{Experiments}
To demonstrate the effectiveness of \method~we present results on both the KITTI 3D object detection benchmark~\cite{Kitti} and the Waymo Open Dataset~\cite{waymo}.

The KITTI 3D object detection benchmark~\cite{Kitti} is divided into 7,481 training samples and 7,518 testing samples. The training samples are commonly divided into a \textit{train} set (3,712 samples) and a
\textit{val} set (3,769 samples) following~\cite{3DOP}, which is also adopted here. We compare \method~with existing methods on the \textit{test} set by training our model on both the \textit{train} and \textit{val} sets. We evaluate on the \textit{val} set for ablation by training our model on only the \textit{train} set.

The Waymo Open Dataset~\cite{waymo} is a more recently released autonomous driving dataset, which consists of 798 training sequences and 202 validation sequences. The dataset also includes 150 test sequences without ground truth data. The dataset provides object labels in the full 360\degree~field of view with a multi-camera rig. We only use the front camera and only consider object labels in the front-camera's field of view (50.4\degree) for the task of monocular object detection, and provide results on the validation sequences. We sample every \nth{3} frame from the training sequences to form our training set (51,564 samples) due to the large dataset size and high frame rate.
\\[0.2\baselineskip]
\noindent
\textbf{Input Parameters}.
The voxel grid is defined by a range and voxel size in 3D space. On KITTI~\cite{Kitti}, we use $[2, 46.8] \times [-30.08, 30.08] \times [-3, 1]$ (\si{\meter}) for the range and $[0.16, 0.16, 0.16]$ (\si{\meter}) for the voxel size for the $x$, $y$, and $z$ axes respectively. On Waymo, we use $[2, 55.76] \times [-25.6, 25.6] \times [-4, 4]$ (\si{\meter}) for the range and $[0.16, 0.16, 0.16]$ (\si{\meter}) for the voxel size. Additionally, we downsample Waymo images to 1248 $\times$ 832.
\\[0.2\baselineskip]
\noindent
\textbf{Training and Inference Details}.
Our method is implemented in PyTorch~\cite{Pytorch}. The network is trained on a NVIDIA Tesla V100 (32G) GPU. The Adam~\cite{Adam} optimizer is used with an initial learning rate of 0.001 and is modified using the one-cycle learning rate policy~\cite{onecycle}. We train the model for 80 epochs on the KITTI dataset~\cite{Kitti} and 10 epochs on the Waymo Open Dataset~\cite{waymo}. We use a batch size of 4 for KITTI~\cite{Kitti} and a batch size of 2 for Waymo.  The values $\lambda_{\mathrm{depth}} = 3.0, \lambda_{\mathrm{cls}} = 1.0, \lambda_{\mathrm{reg}} = 2.0, \lambda_{\mathrm{dir}} = 0.2$ are used for the loss weighting factors in Equation~\ref{eq:total_loss}. We employ horizontal flip as our data augmentation and train one model for all classes. During inference, we filter boxes with a score threshold of 0.1 and apply non-maximum suppression (NMS) with an IoU threshold of 0.01.
\subsection{KITTI Dataset Results} \label{KITTI Dataset Results}
Results on the KITTI dataset~\cite{Kitti} are evaluated using average precision $(\mathrm{AP}|_{R_{40}})$. The evaluation is separated by difficulty settings (Easy, Moderate, and Hard) and by object class (Car, Pedestrian, and Cyclist). The Car class has an IoU criteria of 0.7 while the Pedestrian and Cyclist classes have an IoU criteria of 0.5, where IoU criteria is a threshold to be considered a true positive detection.

Table~\ref{tab:kitti-test} shows the results of \method~on the KITTI~\cite{Kitti} \textit{test} set compared to state-of-the-art published monocular methods, listed in rank order of performance on the Car class at the Moderate difficulty setting. We note that our method outperforms previous single frame methods by large margins on $\mathrm{AP}|_{R_{40}}$ of +2.40\%, +1.69\%, and +1.29\% on the Car class on the Easy, Moderate, and Hard difficulties respectively. Additionally, \method~ranks higher than the multi-frame method Kinematic3D~\cite{Kinematic3D}. Our method also outperforms the previous state-of-the art method on the Pedestrian class MonoPair~\cite{MonoPair} with margins on $\mathrm{AP}|_{R_{40}}$ of +2.85\%, +1.46\%, and +1.23\%. Our method achieves second place on the Cyclist class with margins on $\mathrm{AP}|_{R_{40}}$ of -1.37\%, -1.33\%, and -0.38\% relative to MonoPSR~\cite{monopsr}.
\subsection{Waymo Dataset Results} \label{Waymo Dataset Results}
\begin{table*}
\centering
\resizebox{\textwidth}{!}{%
\begin{tabular}{c|c|cccc|cccc}
\toprule
 &  & \multicolumn{4}{c|}{3D mAP} & \multicolumn{4}{c}{3D mAPH} \\
\multirow{-2}{*}{Difficulty} & \multirow{-2}{*}{Method} & Overall & 0 - 30m & 30 - 50m & 50m - $\infty$ & Overall & 0 - 30m & 30 - 50m & 50m - $\infty$ \\ \hline
 & M3D-RPN~\cite{M3D-RPN} & 0.35 & 1.12 & 0.18 & 0.02 & 0.34 & 1.10 & 0.18 & 0.02 \\
 & \textbf{CaDNN (Ours)} & \textbf{5.03} & \textbf{14.54} & \textbf{1.47} & \textbf{0.10} & \textbf{4.99} & \textbf{14.43} & \textbf{1.45} & \textbf{0.10} \\
\multirow{-3}{*}{\begin{tabular}[c]{@{}c@{}}LEVEL\_1\\ (IOU = 0.7)\end{tabular}} & \cellcolor{TableBlue}\textit{Improvement} & \cellcolor{TableBlue}\textit{+4.69} & \cellcolor{TableBlue}\textit{+13.43} & \cellcolor{TableBlue}\textit{+1.28} & \cellcolor{TableBlue}\textit{+0.08} & \cellcolor{TableBlue}\textit{+4.65} & \cellcolor{TableBlue}\textit{+13.33} & \cellcolor{TableBlue}\textit{+1.28} & \cellcolor{TableBlue}\textit{+0.08} \\ \hline
 & M3D-RPN~\cite{M3D-RPN} & {\color[HTML]{1D1C1D} 0.33} & {\color[HTML]{1D1C1D} 1.12} & {\color[HTML]{1D1C1D} 0.18} & {\color[HTML]{1D1C1D} 0.02} & {\color[HTML]{1D1C1D} 0.33} & {\color[HTML]{1D1C1D} 1.10} & {\color[HTML]{1D1C1D} 0.17} & {\color[HTML]{1D1C1D} 0.02} \\
 & \textbf{CaDNN (Ours)} & \textbf{4.49} & \textbf{14.50} & \textbf{1.42} & \textbf{0.09} & \textbf{4.45} & \textbf{14.38} & \textbf{1.41} & \textbf{0.09} \\
\multirow{-3}{*}{\begin{tabular}[c]{@{}c@{}}LEVEL\_2\\ (IOU = 0.7)\end{tabular}} & \cellcolor{TableBlue}\textit{Improvement} & \cellcolor{TableBlue}\textit{+4.15} & \cellcolor{TableBlue}\textit{+13.38} & \cellcolor{TableBlue}\textit{+1.24} & \cellcolor{TableBlue}\textit{+0.07} & \cellcolor{TableBlue}\textit{+4.12} & \cellcolor{TableBlue}\textit{+13.28} & \cellcolor{TableBlue}\textit{+1.24} & \cellcolor{TableBlue}\textit{+0.07} \\ \hline
 & M3D-RPN~\cite{M3D-RPN} & 3.79 & 11.14 & 2.16 & 0.26 & 3.63 & 10.70 & 2.09 & 0.21 \\
 & \textbf{CaDNN (Ours)} & \textbf{17.54} & \textbf{45.00} & \textbf{9.24} & \textbf{0.64} & \textbf{17.31} & \textbf{44.46} & \textbf{9.11} & \textbf{0.62} \\
\multirow{-3}{*}{\begin{tabular}[c]{@{}c@{}}LEVEL\_1\\ (IOU = 0.5)\end{tabular}} & \cellcolor{TableBlue}\textit{Improvement} & \cellcolor{TableBlue}\textit{+13.76} & \cellcolor{TableBlue}\textit{+33.86} & \cellcolor{TableBlue}\textit{+7.08} & \cellcolor{TableBlue}\textit{+0.39} & \cellcolor{TableBlue}\textit{+13.69} & \cellcolor{TableBlue}\textit{+33.77} & \cellcolor{TableBlue}\textit{+7.02} & \cellcolor{TableBlue}\textit{+0.41} \\ \hline
 & M3D-RPN~\cite{M3D-RPN} & 3.61 & 11.12 & 2.12 & 0.24 & 3.46 & 10.67 & 2.04 & 0.20 \\
 & \textbf{CaDNN (Ours)} & \textbf{16.51} & \textbf{44.87} & \textbf{8.99} & \textbf{0.58} & \textbf{16.28} & \textbf{44.33} & \textbf{8.86} & \textbf{0.55} \\
\multirow{-3}{*}{\begin{tabular}[c]{@{}c@{}}LEVEL\_2\\ (IOU = 0.5)\end{tabular}} & \cellcolor{TableBlue}\textit{Improvement} & \cellcolor{TableBlue}\textit{+12.89} & \cellcolor{TableBlue}\textit{+33.75} & \cellcolor{TableBlue}\textit{+6.87} & \cellcolor{TableBlue}\textit{+0.34} & \cellcolor{TableBlue}\textit{+12.82} & \cellcolor{TableBlue}\textit{+33.66} & \cellcolor{TableBlue}\textit{+6.81} & \cellcolor{TableBlue}\textit{+0.36} \\ \bottomrule
\end{tabular}
}
\caption{Results on the Waymo Open Dataset Validation Set on the Vehicle class. We evaluate M3D-RPN~\cite{M3D-RPN} as a baseline for comparison.}
\label{tab:waymo-val}
\end{table*}
We adopt the officially released evaluation to calculate the mean average precision (mAP) and the mean average precision weighted by heading (mAPH) on the Waymo Open Dataset~\cite{waymo}. The evaluation is separated by difficulty setting (LEVEL\_1, LEVEL\_2) and distance to the sensor (0 - 30\si{\meter}, 30 - 50\si{\meter}, and 50\si{\meter} - $\infty$). We evaluate on the Vehicle class with an IoU criteria of 0.7 and 0.5.

To the best of our knowledge, no monocular methods have reported results on Waymo. In order to provide a baseline, we extend the official implementation of M3D-RPN~\cite{M3D-RPN} to support the Waymo Open Dataset~\cite{waymo}. Table~\ref{tab:waymo-val} shows the results of both the M3D-RPN~\cite{M3D-RPN} baseline and \method~on the Waymo validation set. Our method significantly outperforms M3D-RPN~\cite{M3D-RPN} with margins on AP/APH of +4.69\%/+4.65\% and +4.15\%/+4.12\% on the LEVEL\_1 and LEVEL\_2 difficulties respectively for an IoU criteria of 0.7.
\subsection{Ablation Studies} \label{sec:Ablation Studies}
\begin{table}
\small
\centering
\begin{tabular}{x{0.6cm}|cccc|ccc}
\toprule
\multirow{2}{*}{Exp.} & \multirow{2}{*}{$\mathbf{D}$} & \multirow{2}{*}{$L_{\mathrm{depth}}$} & \multirow{2}{*}{$\alpha_{\mathrm{fg}}$} & \multirow{2}{*}{LID} & \multicolumn{3}{c}{Car (IOU = 0.7)} \\
 &  &  &  &  & Easy & Mod. & Hard \\ \hline
\newtag{1}{itm:ab1} &  &  &  &  & 7.83 & 5.66 & 4.84 \\
\newtag{2}{itm:ab2} & \checkmark &  &  & & 9.33 & 6.43 & 5.30 \\
\newtag{3}{itm:ab3} & \checkmark & \checkmark &  & & 19.73 & 14.03 & 11.84 \\
\newtag{4}{itm:ab4} & \checkmark & \checkmark & \checkmark &  & 20.40 & 15.10 & 12.75 \\
\newtag{5}{itm:ab5} & \checkmark & \checkmark & \checkmark & \checkmark & \textbf{23.57} & \textbf{16.31} & \textbf{13.84} \\
\bottomrule
\end{tabular}
\caption{\method~Ablation Experiments on the KITTI \textit{val} set using $\left.\mathrm{AP}\right|_{R_{40}}$. $\mathbf{D}$ indicates depth distribution prediction, $L_{\mathrm{depth}}$ indicates depth distribution supervision. $\alpha_{\mathrm{fg}}$ indicates separate setting of loss weighting factor for foreground object pixels in the depth loss function $L_{\mathrm{depth}}$. LID indicates the LID discretization method.}
\label{tab:ablate}
\bigskip
\begin{tabular}{x{0.59cm}|ccc|ccc}
\toprule
\multirow{2}{*}{Exp.} & \multirow{2}{*}{$\mathbf{D}$} & \multirow{2}{*}{$L_{\mathrm{depth}}$} & \multirow{2}{*}{$\otimes$} & \multicolumn{3}{c}{Car (IOU = 0.7)} \\
 &  &  &  &  Easy & Mod. & Hard \\ \hline
\newtag{1}{itm:dp1} & BTS~\cite{BTS} & Sep. &  & 16.69 & 10.18 & 8.63 \\
\newtag{2}{itm:dp2} & DORN~\cite{DORN} & Sep. & & 16.43 & 11.04 & 9.65 \\
\newtag{3}{itm:dp3} & \method & Sep. &  & 17.64 & 12.26 & 10.10 \\
\newtag{4}{itm:dp4} & \method & Joint & & 20.61 & 13.71 & 11.96 \\
\newtag{5}{itm:dp5} & \method & Joint & \checkmark & \textbf{23.57} & \textbf{16.31} & \textbf{13.84} \\ \bottomrule
\end{tabular}
\caption{\method~Depth Estimation Ablation on the KITTI \textit{val} set using $\left.\mathrm{AP}\right|_{R_{40}}$. $\mathbf{D}$ indicates the source of the depth estimates used to generate depth distributions. $L_{\mathrm{depth}}$ indicates if depth estimation and object detection are seperately or jointly optimized. $\otimes$ indicates if full distributions are used to generate frustum features $\mathbf{G}$.}
\label{tab:ablate-depth-estimation}
\end{table}
We provide ablation studies on our network to validate our design choices. The results are shown in Tables~\ref{tab:ablate} and~\ref{tab:ablate-depth-estimation}.
\\[0.2\baselineskip]
\noindent
\textbf{Sharpness in Depth Distributions}.
Experiment~\ref{itm:ab1} in Table~\ref{tab:ablate} shows the detection performance when frustum features $\mathbf{G}$ are populated by repeating image features $\mathbf{F}$ along depth axis $d_i$. Experiment~\ref{itm:ab2} adds depth distribution predictions $\mathbf{D}$ to separately weigh image features $\mathbf{F}$, which improves performance on $\mathrm{AP}|_{R_{40}}$ by +1.50\%, +0.77\%, and +0.46\% on the Car class on the Easy, Moderate, and Hard difficulties respectively. Performance is greatly increased (+10.40\%, +7.60\%, +6.54\%) once depth distribution supervision is added in Experiment~\ref{itm:ab3} validating its inclusion. The addition of depth distribution supervision encourages sharp and accurate categorical depth distributions, that encourages image information to be located in 3D space where depth estimation is both accurate and confident. Encouraging sharpness around correct depth bins results in object features that are uniquely located and easily distinguished (see Figure~\ref{fig:intro}) in the BEV projection.
\\[0.2\baselineskip]
\noindent
\textbf{Object Weighting for Depth Distribution Estimation}.
Experiments~\ref{itm:ab1},~\ref{itm:ab2}, and~\ref{itm:ab3} in Table~\ref{tab:ablate} use a fixed loss weighting factor $\alpha = 0.25$ for all pixels in the depth loss function $L_{\mathrm{depth}}$. Experiment~\ref{itm:ab4} shows an improvement (+0.67\%, +1.07\%, +0.91\%) after depth loss weights $\alpha_{\mathrm{fg}} = 3.25/\alpha_{\mathrm{bg}} = 0.25$ are set seperately for foreground object and background pixels (see Section~\ref{sec:TrainingLosses}). Setting a larger foreground object weighting factor $\alpha_{\mathrm{fg}}$ encourages depth estimation to be prioritized for object pixels, leading to more accurate depth estimation and localization for objects.
\\[0.2\baselineskip]
\noindent
\textbf{Linear Increasing Discretization}.
Experiment~\ref{itm:ab5} in Table~\ref{tab:ablate} shows the detection performance improvement (+3.17\%, +1.21\%, +1.09\%) when LID (see Section~\ref{sec:Depth Discretization}) is used rather than uniform discretization. We attribute the performance increase to the accurate depth estimation LID provides across all depths~\cite{Center3D}.
\\[0.2\baselineskip]
\noindent
\textbf{Joint Depth Understanding}.
Experiments~\ref{itm:dp1},~\ref{itm:dp2} and~\ref{itm:dp3} in Table~\ref{tab:ablate-depth-estimation} show the detection performance with separate depth estimation from BTS~\cite{BTS}, DORN~\cite{DORN}, and \method~respectively. The depth maps from BTS~\cite{BTS} and DORN~\cite{DORN} are converted to depth bin indices using LID discretization as outlined in Section~\ref{sec:Depth Discretization}, and converted to a one-hot encoding to generate the depth distributions $\mathbf{D}$. The one-hot encoding places the image feature at a single depth bin indicated by the input depth map when generating frustum features $\mathbf{G}$. We constuct an equivalent version of \method~that selects a single depth bin for each pixel, by selecting the bin with highest probability for each distribution in $\mathbf{D}$. Experiment~\ref{itm:dp4} shows improved performance (+2.97\%, +1.45\%, +1.86\%) when depth estimation and object detection are performed jointly, which we attribute to the well-known benefits of end-to-end learning for 3D detection.
\\[0.2\baselineskip]
\noindent
\textbf{Categorical Depth Distributions}.
Experiment~\ref{itm:ab5} in Table~\ref{tab:ablate-depth-estimation} uses the full depth distribution $\mathbf{D}$ in the frustum features computation $\mathbf{G}=\mathbf{D}\otimes\mathbf{F}$, leading to a clear increase in performance (+2.96\%, 2.60\%, 1.88\%). We attribute the performance increase to the additional depth uncertainty information embedded in the feature representations.
\subsection{Depth Distribution Uncertainty} \label{sec:Depth Distribution Uncertainty}
\begin{figure}
\begin{center}
\includesvg[width=\columnwidth]{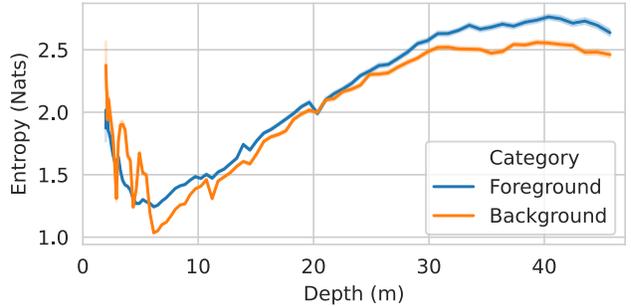}
\end{center}
\vspace{-3mm}
\caption{We plot the entropy of the estimated depth distributions $\mathbf{D}$ against depth. We show both the mean (solid line) and 95\% confidence interval (shaded region) at each ground truth depth bin.}
\label{fig:entropy}
\end{figure}
To validate that our depth distributions contain meaningful uncertainty information, we compute the Shanon entropy for each estimated categorical depth distribution in $\mathbf{D}$. We label each distribution with its associated ground truth depth bin and foreground/background classification. For each group, we compute the entropy statistics which are shown in Figure~\ref{fig:entropy}. We observe that entropy generally increases as a function of depth, where depth estimates are challenging, indicating our distributions describe meaningful uncertainty information. Our network produces the lowest distribution entropy at pixels with ground truth depth of around 6 meters. We attribute the high entropy at depths closer than 6 meters to the small number of pixels at shorter ranges in the training set. Finally, we note that the foreground depth distribution estimates have slightly higher entropy than background pixels, a phenomenon that can also be attributed to training set imbalance.

\section{Conclusion} \label{Conclusions}
We have presented \method, a novel monocular 3D object detection method that estimates accurate categorical depth distributions for each pixel. The depth distributions are combined with the image features to generate bird's-eye-view representations that retain depth confidence, to be exploited for 3D object detection. We have shown that estimating sharp categorical distributions centered around the correct depth value, and jointly performing depth estimation and object detection is vital for 3D object detection performance, leading to a \nth{1} place ranking on the KITTI dataset~\cite{KITTI_Test} among all published methods at the time of submission.
{\small
\bibliographystyle{ieee_fullname}
\bibliography{egbib}

\begin{thebibliography}{10}\itemsep=-1pt

\bibitem{KITTI_Test}
Kitti's 3d object detection evaluation benchmark 2017.
\newblock
  \url{http://www.cvlibs.net/datasets/kitti/eval_object.php?obj_benchmark=3d}{}.
  Accessed on 15.11.2020.

\bibitem{DiffRendering}
Deniz Beker, Hiroharu Kato, Mihai~Adrian Morariu, Takahiro Ando, Toru Matsuoka,
  Wadim Kehl, and Adrien Gaidon.
\newblock Monocular differentiable rendering for self-supervised 3d object
  detection.
\newblock {\em ECCV}, 2020.

\bibitem{M3D-RPN}
Garrick Brazil and Xiaoming Liu.
\newblock {M3D-RPN:} monocular {3D} region proposal network for object
  detection.
\newblock {\em ICCV}, 2019.

\bibitem{Kinematic3D}
Garrick Brazil, Gerard Pons-Moll, Xiaoming Liu, and Bernt Schiele.
\newblock Kinematic 3d object detection in monocular video.
\newblock {\em ECCV}, 2020.

\bibitem{DeepMANTA}
Florian Chabot, Mohamed Chaouch, Jaonary Rabarisoa, C{\'{e}}line
  Teuli{\`{e}}re, and Thierry Chateau.
\newblock Deep {MANTA:} {A} coarse-to-fine many-task network for joint 2d and
  {3D} vehicle analysis from monocular image.
\newblock {\em CVPR}, 2017.

\bibitem{DeepLabV3}
Liang{-}Chieh Chen, George Papandreou, Florian Schroff, and Hartwig Adam.
\newblock Rethinking atrous convolution for semantic image segmentation.
\newblock {\em arXiv preprint}, 2017.

\bibitem{DeepLabAlt}
Liang-Chieh Chen, George Papandreou, Iasonas Kokkinos, Kevin Murphy, and
  Alan~L. Yuille.
\newblock Semantic image segmentation with deep convolutional nets and fully
  connected crfs.
\newblock {\em ICLR}, 2015.

\bibitem{DeepLab}
Liang-Chieh Chen, George Papandreou, Iasonas Kokkinos, Kevin Murphy, and
  Alan~L. Yuille.
\newblock Deeplab: Semantic image segmentation with deep convolutional nets,
  atrous convolution, and fully connected crfs.
\newblock {\em arXiv preprint}, 2016.

\bibitem{Mono3D}
Xiaozhi Chen, Kaustav Kundu, Ziyu Zhang, Huimin Ma1, Sanja Fidler, and Raquel
  Urtasun.
\newblock Monocular 3d object detection for autonomous driving.
\newblock {\em CVPR}, 2016.

\bibitem{3DOP}
Xiaozhi Chen, Kaustav Kundu, Yukun Zhu, Andrew~G Berneshawi, Huimin Ma, Sanja
  Fidler, and Raquel Urtasun.
\newblock 3d object proposals for accurate object class detection.
\newblock {\em NIPS}, 2015.

\bibitem{DSGN}
Yilun Chen, Shu Liu, Xiaoyong Shen, and Jiaya Jia.
\newblock Dsgn: Deep stereo geometry network for 3d object detection.
\newblock {\em CVPR}, 2020.

\bibitem{MonoPair}
Yongjian Chen, Lei Tai, Kai Sun, and Mingyang Li.
\newblock Monopair: Monocular 3d object detection using pairwise spatial
  relationships.
\newblock {\em CVPR}, 2020.

\bibitem{D4LCN}
Mingyu Ding, Yuqi Huo, Hongwei Yi, Zhe Wang, Jianping Shi, Zhiwu Lu, and Ping
  Luo.
\newblock Learning depth-guided convolutions for monocular 3d object detection.
\newblock {\em CVPR}, 2020.

\bibitem{Eigen2}
David Eigen and Rob Fergus.
\newblock Predicting depth, surface normals and semantic labels with a common
  multi-scale convolutional architecture.
\newblock {\em ICCV}, 2015.

\bibitem{DORN}
Huan Fu, Mingming Gong, Chaohui Wang, Kayhan Batmanghelich, and Dacheng Tao.
\newblock Deep ordinal regression network for monocular depth estimation.
\newblock {\em CVPR}, 2018.

\bibitem{Kitti}
Andreas Geiger, Philip Lenz, and Raquel Urtasun.
\newblock Are we ready for autonomous driving? the kitti vision benchmark
  suite.
\newblock {\em CVPR}, 2012.

\bibitem{ResNet}
Kaiming He, Xiangyu Zhang, Shaoqing Ren, and Jian Sun.
\newblock Deep residual learning for image recognition.
\newblock {\em CVPR}, 2016.

\bibitem{Adam}
Diederik~P Kingma and Jimmy Ba.
\newblock Adam: A method for stochastic optimization.
\newblock {\em ICLR}, 2015.

\bibitem{Class}
S.~B. Kotsiantis.
\newblock Supervised machine learning: A review of classification techniques.
\newblock In {\em Proceedings of the 2007 Conference on Emerging Artificial
  Intelligence Applications in Computer Engineering: Real Word AI Systems with
  Applications in EHealth, HCI, Information Retrieval and Pervasive
  Technologies}, NLD, 2007. IOS Press.

\bibitem{IPBasic}
Jason Ku, Ali Harakeh, and Steven~Lake Waslander.
\newblock In defense of classical image processing: Fast depth completion on
  the {CPU}.
\newblock {\em CRV}, 2018.

\bibitem{AVOD}
Jason Ku, Melissa Mozifian, Jungwook Lee, Ali Harakeh, and Steven~Lake
  Waslander.
\newblock Joint {3D} proposal generation and object detection from view
  aggregation.
\newblock {\em IROS}, 2018.

\bibitem{monopsr}
Jason Ku, Alex~D. Pon, and Steven~L. Waslander.
\newblock Monocular {3D} object detection leveraging accurate proposals and
  shape reconstruction.
\newblock {\em CVPR}, 2019.

\bibitem{Sharpness_Proof}
Volodymyr Kuleshov, Nathan Fenner, and Stefano Ermon.
\newblock Accurate uncertainties for deep learning using calibrated regression.
\newblock {\em arXiv preprint arXiv:1807.00263}, 2018.

\bibitem{3DRCNN}
Abhijit Kundu, Yin Li, and James~M. Rehg.
\newblock {3D-RCNN}: Instance-level {3D} object reconstruction via
  render-and-compare.
\newblock {\em CVPR}, 2018.

\bibitem{Laina}
Iro Laina, Christian Rupprecht, Vasileios Belagiannis, Federico Tombari, and
  Nassir Navab.
\newblock Deeper depth prediction with fully convolutional residual networks.
\newblock {\em 3DV}, 2016.

\bibitem{PointPillars}
Alex~H. Lang, Sourabh Vora, Holger Caesar, Lubing Zhou, Jiong Yang, and Oscar
  Beijbom.
\newblock {PointPillars}: Fast encoders for object detection from point clouds.
\newblock {\em CVPR}, 2019.

\bibitem{BTS}
Jin~Han Lee, Myung{-}Kyu Han, Dong~Wook Ko, and Il~Hong Suh.
\newblock From big to small: Multi-scale local planar guidance for monocular
  depth estimation.
\newblock {\em arXiv preprint}, 2019.

\bibitem{CGStereo}
Chengyao Li, Jason Ku, and Steven~L. Waslander.
\newblock Confidence guided stereo 3d object detection with split depth
  estimation.
\newblock {\em IROS}, 2020.

\bibitem{RTM3D}
Peixuan Li, Huaici Zhao, Pengfei Liu, and Feidao Cao.
\newblock Rtm3d: Real-time monocular 3d detection from object keypoints for
  autonomous driving.
\newblock {\em ECCV}, 2020.

\bibitem{FocalLoss}
Tsung{-}Yi Lin, Priya Goyal, Ross~B. Girshick, Kaiming He, and Piotr
  Doll{\'{a}}r.
\newblock Focal loss for dense object detection.
\newblock {\em PAMI}, 2018.

\bibitem{COCO}
Tsung{-}Yi Lin, Michael Maire, Serge~J. Belongie, Lubomir~D. Bourdev, Ross~B.
  Girshick, James Hays, Pietro Perona, Deva Ramanan, Piotr Doll{\'{a}}r, and
  C.~Lawrence Zitnick.
\newblock Microsoft {COCO:} common objects in context.
\newblock {\em ECCV}, 2014.

\bibitem{RARNet}
Lijie Liu, Chufan Wu, Jiwen Lu, Lingxi Xie, Jie Zhou, and Qi Tian.
\newblock Reinforced axial refinement network for monocular 3d object
  detection.
\newblock {\em ECCV}, 2020.

\bibitem{SMOKE}
Zechen Liu, Zizhang Wu, and Roland Toth.
\newblock Smoke: Single-stage monocular 3d object detection via keypoint
  estimation.
\newblock {\em CVPRW}, 2020.

\bibitem{FCN}
Jonathan Long, Evan Shelhamer, and Trevor Darrell.
\newblock Fully convolutional networks for semantic segmentation.
\newblock {\em CVPR}, 2015.

\bibitem{VAEMap}
Chenyang Lu, Gijs Dubbelman, and Marinus Jacobus~Gerardus van~de Molengraft.
\newblock Monocular semantic occupancy grid mapping with convolutional
  variational auto-encoders.
\newblock {\em ICRA}, 2019.

\bibitem{PatchNet}
Xinzhu Ma, Shinan Liu, Zhiyi Xia, Hongwen Zhang, Xingyu Zeng, and Wanli Ouyang.
\newblock Rethinking pseudo-lidar representation.
\newblock {\em ECCV}, 2020.

\bibitem{AM3D}
Xinzhu Ma, Zhihui Wang, Haojie Li, Wanli Ouyang, and Pengbo Zhang.
\newblock Accurate monocular {3D} object detection via color-embedded {3D}
  reconstruction for autonomous driving.
\newblock {\em ICCV}, 2019.

\bibitem{ROI-10D}
Fabian Manhardt, Wadim Kehl, and Adrien Gaidon.
\newblock {ROI-10D:} monocular lifting of 2d detection to 6d pose and metric
  shape.
\newblock {\em CVPR}, 2019.

\bibitem{MonoLayout}
Kaustubh Mani, Swapnil Daga, Shubhika Garg, Sai~Shankar Narasimhan, Madhava
  Krishna, and Krishna~Murthy Jatavallabhula.
\newblock Monolayout: Amodal scene layout from a single image.
\newblock {\em WACV}, 2020.

\bibitem{Deep3DBox}
Arsalan Mousavian, Dragomir Anguelov, John Flynn, and Jana Kosecka.
\newblock 3d bounding box estimation using deep learning and geometry.
\newblock {\em CVPR}, 2016.

\bibitem{bevseg}
Mong~H. Ng, Kaahan Radia, Jianfei Chen, Dequan Wang, Ionel Gog, and Joseph~E.
  Gonzalez.
\newblock Bev-seg: Bird's eye view semantic segmentation using geometry and
  semantic point cloud.
\newblock {\em CVPRW}, 2020.

\bibitem{CrossView}
Bowen Pan, Jiankai Sun, Alex Andonian, Aude Oliva, and Bolei Zhou.
\newblock Cross-view semantic segmentation for sensing surroundings.
\newblock {\em ICRA}, 2019.

\bibitem{Pytorch}
Adam Paszke, Sam Gross, Francisco Massa, Adam Lerer, James Bradbury, Gregory
  Chanan, Trevor Killeen, Zeming Lin, Natalia Gimelshein, Luca Antiga, Alban
  Desmaison, Andreas Kopf, Edward Yang, Zachary DeVito, Martin Raison, Alykhan
  Tejani, Sasank Chilamkurthy, Benoit Steiner, Lu Fang, Junjie Bai, and Soumith
  Chintala.
\newblock Pytorch: An imperative style, high-performance deep learning library.
\newblock In {\em NeurIPS}. Curran Associates, Inc., 2019.

\bibitem{LiftSplatShoot}
Jonah Philion and Sanja Fidler.
\newblock Lift, splat, shoot: Encoding images from arbitrary camera rigs by
  implicitly unprojecting to 3d.
\newblock {\em ECCV}, 2020.

\bibitem{OCStereo}
Alex~D. Pon, Jason Ku, Chengyao Li, and Steven~L. Waslander.
\newblock Object-centric stereo matching for 3d object detection.
\newblock {\em ICRA}, 2020.

\bibitem{e2e_pseudo-lidar}
Rui Qian, Divyansh Garg, Yan Wang, Yurong You, Serge Belongie, Bharath
  Hariharan, Mark Campbell, Kilian~Q. Weinberger, and Wei-Lun Chao.
\newblock End-to-end pseudo-lidar for image-based 3d object detection.
\newblock {\em CVPR}, 2020.

\bibitem{PON}
Thomas Roddick and Roberto Cipolla.
\newblock Predicting semantic map representations from images using pyramid
  occupancy networks.
\newblock {\em CVPR}, 2020.

\bibitem{OFT}
Thomas Roddick, Alex Kendall, and Roberto Cipolla.
\newblock Orthographic feature transform for monocular {3D} object detection.
\newblock {\em BMVC}, 2018.

\bibitem{Learning}
Samuel Schulter, Menghua Zhai, Nathan Jacobs, and Manmohan Chandraker.
\newblock Learning to look around objects for top-view representations of
  outdoor scenes.
\newblock {\em ECCV}, 2018.

\bibitem{PV-RCNN}
Shaoshuai Shi, Chaoxu Guo, Li Jiang, Zhe Wang, Jianping Shi, Xiaogang Wang, and
  Hongsheng Li.
\newblock {PV-RCNN}: Point-voxel feature set abstraction for {3D} object
  detection.
\newblock {\em CVPR}, 2020.

\bibitem{PointRCNN}
Shaoshuai Shi, Xiaogang Wang, and Hongsheng Li.
\newblock {PointRCNN}: {3D} object proposal generation and detection from point
  cloud.
\newblock {\em CVPR}, 2019.

\bibitem{MonoDis}
Andrea Simonelli, Samuel~Rota Bul{\`{o}}, Lorenzo Porzi, Manuel
  L{\'{o}}pez{-}Antequera, and Peter Kontschieder.
\newblock Disentangling monocular 3d object detection.
\newblock {\em ICCV}, 2019.

\bibitem{MoVi-3D}
Andrea Simonelli, Samuel~Rota Bulò, Lorenzo Porzi, Elisa Ricci, and Peter
  Kontschieder.
\newblock Towards generalization across depth for monocular 3d object
  detection.
\newblock {\em ECCV}, 2020.

\bibitem{onecycle}
Leslie~N. Smith.
\newblock A disciplined approach to neural network hyper-parameters: Part 1 -
  learning rate, batch size, momentum, and weight decay.
\newblock {\em arXiv preprint}, 2018.

\bibitem{BirdGAN}
Siddharth Srivastava, Fr{\'{e}}d{\'{e}}ric Jurie, and Gaurav Sharma.
\newblock Learning 2d to 3d lifting for object detection in 3d for autonomous
  vehicles.
\newblock {\em arXiv preprint}, 2019.

\bibitem{waymo}
Pei Sun, Henrik Kretzschmar, Xerxes Dotiwalla, Aurelien Chouard, Vijaysai
  Patnaik, Paul Tsui, James Guo, Yin Zhou, Yuning Chai, Benjamin Caine, Vijay
  Vasudevan, Wei Han, Jiquan Ngiam, Hang Zhao, Aleksei Timofeev, Scott
  Ettinger, Maxim Krivokon, Amy Gao, Aditya Joshi, Yu Zhang, Jonathon Shlens,
  Zhifeng Chen, and Dragomir Anguelov.
\newblock Scalability in perception for autonomous driving: Waymo open dataset,
  2019.

\bibitem{Center3D}
Yunlei Tang, Sebastian Dorn, and Chiragkumar Savani.
\newblock Center3d: Center-based monocular 3d object detection with joint depth
  understanding.
\newblock {\em arXiv preprint}, 2020.

\bibitem{SDC-Depth}
Lijun Wang, Jianming Zhang, Oliver Wang, Zhe Lin, and Huchuan Lu.
\newblock Sdc-depth: Semantic divide-and-conquer network for monocular depth
  estimation.
\newblock {\em CVPR}, 2020.

\bibitem{Psuedo-LIDAR}
Yan Wang, Wei{-}Lun Chao, Divyansh Garg, Bharath Hariharan, Mark Campbell, and
  Kilian~Q. Weinberger.
\newblock {Pseudo-LiDAR} from visual depth estimation: Bridging the gap in {3D}
  object detection for autonomous driving.
\newblock {\em CVPR}, 2019.

\bibitem{TopView}
Ziyan Wang, Buyu Liu, Samuel Schulter, and Manmohan Chandraker.
\newblock A parametric top-view representation of complex road scenes.
\newblock {\em CVPR}, 2019.

\bibitem{Mono3D_Plidar}
Xinshuo Weng and Kris Kitani.
\newblock Monocular 3d object detection with pseudo-lidar point cloud.
\newblock {\em ICCVW}, 2019.

\bibitem{MultiFusion}
Bin Xu and Zhenzhong Chen.
\newblock Multi-level fusion based {3D} object detection from monocular images.
\newblock {\em CVPR}, 2018.

\bibitem{PAD-Net}
Dan Xu, Wanli Ouyang, Xiaogang Wang, and Nicu Sebe.
\newblock Pad-net: Multi-tasks guided prediction-and-distillation network for
  simultaneous depth estimation and scene parsing.
\newblock {\em CVPR}, 2018.

\bibitem{UR3D}
Tae-Kyun~Kim Xuepeng~Shi, Zhixiang~Chen.
\newblock Distance-normalized unified representation for monocular 3d object
  detection.
\newblock {\em ECCV}, 2020.

\bibitem{DA-3Ddet}
Xiaoqing Ye, Liang Du, Yifeng Shi, Yingying Li, Xiao Tan, Jianfeng Feng, Errui
  Ding, and Shilei Wen.
\newblock Monocular 3d object detection via feature domain adaptation.
\newblock {\em ECCV}, 2020.

\bibitem{JointSeg}
Zhenyu Zhang, Zhen Cui, Chunyan Xu, Zequn Jie, Xiang Li, and Jian Yang.
\newblock Joint task-recursive learning for semantic segmentation and depth
  estimation.
\newblock {\em ECCV}, 2018.

\end{thebibliography}
}
\appendix
\section{Additional Results} \label{sec:AddResults}
\subsection{KITTI Dataset Results} \label{Add KITTI Dataset Results}
\begin{table*}
\centering
\begin{tabular}{p{3.6cm}|c|x{0.9cm}x{0.9cm}x{0.9cm}|x{0.9cm}x{0.9cm}x{0.9cm}|x{0.9cm}x{0.9cm}x{0.9cm}}
\toprule
 &  & \multicolumn{3}{c|}{Car (IOU = 0.7)} & \multicolumn{3}{c|}{Pedestrian (IOU = 0.5)} & \multicolumn{3}{c}{Cyclist (IOU = 0.5)} \\
\multirow{-2}{*}{Method} & \multirow{-2}{*}{Frames} & Easy & \textbf{Mod.} & Hard & Easy & \textbf{Mod.} & Hard & Easy & \textbf{Mod.} & Hard \\ \hline
Kinematic3D~\cite{Kinematic3D} & 4 & 26.69 & 17.52 & 13.10 & -- & -- & -- & -- & -- & -- \\ \hline
ROI-10D~\cite{ROI-10D} & 1 & 9.78 & 4.91 & 3.74 & -- & -- & -- & -- & -- & -- \\
UR3D~\cite{UR3D} & 1 & 21.85 & 12.51 & 9.20 & -- & -- & -- & -- & -- & -- \\
MonoPSR~\cite{monopsr} & 1 & 18.33 & 12.58 & 9.91 & 7.24 & 4.56 & 4.11 & {\color[HTML]{FE0000} \textbf{9.87}} & {\color[HTML]{FE0000} \textbf{5.78}} & {\color[HTML]{3531FF} \textbf{4.57}} \\
MonoDIS~\cite{MonoDis} & 1 & 17.23 & 13.19 & 11.12 & -- & -- & -- & -- & -- & -- \\
M3D-RPN~\cite{M3D-RPN} & 1 & 21.02 & 13.67 & 10.23 & 5.65 & 4.05 & 3.29 & 1.25 & 0.81 &  0.78 \\
Mono3D-PLiDAR~\cite{Mono3D_Plidar} & 1 & 21.27 & 13.92 & 11.25 & -- & -- & -- & -- & -- & -- \\
RTM3D~\cite{RTM3D} & 1 & 19.17 & 14.20 & 11.99 & -- & -- & -- & -- & -- & -- \\
SMOKE~\cite{SMOKE} & 1 & 20.83 & 14.49 & 12.75 & -- & -- & -- & -- & -- & -- \\
MonoPair~\cite{MonoPair} & 1 & 19.28 & 14.83 & 12.89 & {\color[HTML]{3531FF} \textbf{10.99}} & {\color[HTML]{3531FF} \textbf{7.04}} &  {\color[HTML]{3531FF} \textbf{6.29}} & 4.76 & 2.87 & 2.42  \\
RAR-Net~\cite{RARNet} & 1 & 22.45 & 15.02 & 12.93 & -- & -- & -- & -- & -- & -- \\
D4LCN~\cite{D4LCN} & 1 & 22.51 & 16.02 & 12.55  & 5.06 & 3.86 &  3.59 & 2.72 & 1.82 & 1.79 \\
PatchNet~\cite{PatchNet} & 1 & 22.97 &  16.86 & {\color[HTML]{3531FF} \textbf{14.97}} & -- & -- & -- & -- & -- & -- \\
MoVi-3D~\cite{MoVi-3D} & 1 & 22.76 & 17.03 & 14.85 & 10.08 & 6.29 & 5.37 & 1.45 & 0.91 & 0.93 \\
AM3D~\cite{AM3D} & 1 & {\color[HTML]{3531FF} \textbf{25.03}} & {\color[HTML]{3531FF} \textbf{17.32}} & 14.91 & -- & -- & -- & -- & -- & -- \\ \hline
\textbf{\method~(ours)} & 1 & {\color[HTML]{FE0000} \textbf{27.94}} & {\color[HTML]{FE0000} \textbf{18.91}} & {\color[HTML]{FE0000} \textbf{17.19}} & {\color[HTML]{FE0000} \textbf{14.72}} & {\color[HTML]{FE0000} \textbf{9.41}} & {\color[HTML]{FE0000} \textbf{8.17}} & {\color[HTML]{3531FF} \textbf{9.67}} & {\color[HTML]{3531FF} \textbf{5.38}} & {\color[HTML]{FE0000} \textbf{4.75}} \\
\cellcolor{TableBlue}\textit{Improvement} & \cellcolor{TableBlue}-- & \cellcolor{TableBlue}\textit{+2.91} & \cellcolor{TableBlue}\textit{+1.59} & \cellcolor{TableBlue}\textit{+2.22} & \cellcolor{TableBlue}\textit{+3.73} & \cellcolor{TableBlue}\textit{+2.37} & \cellcolor{TableBlue}\textit{+1.88} & \cellcolor{TableBlue}\textit{-0.20} & \cellcolor{TableBlue}\textit{-0.40} & \cellcolor{TableBlue}\textit{+0.18} \\ \bottomrule
\end{tabular}
\caption{BEV detection results on the KITTI~\cite{Kitti} \textit{test} set. Results are shown using the $\mathrm{AP}|_{R_{40}}$ metric only for results that are readily available. We indicate the highest result with {\color[HTML]{FE0000} \textbf{red}} and the second highest with {\color[HTML]{3531FF} \textbf{blue}}. Full results for \method~can be accessed \href{http://www.cvlibs.net/datasets/kitti/eval_object_detail.php?&result=d593e0abe5e5f8b0a0ea1523816e7851b0266cd8}{here}.}
\label{tab:kitti-test-bev}
\end{table*}
Table~\ref{tab:kitti-test-bev} shows the results of \method~on the KITTI~\cite{Kitti} \textit{test} set for BEV detection. Our method outpeforms previous single frame methods by large margins on $\mathrm{AP}|_{R_{40}}$ of +2.91\%, +1.59\%, and +2.22\% on the Car class on the Easy, Moderate, and Hard difficulties respectively. Our method outperforms the previous state-of-the art method on the Pedestrian class MonoPair~\cite{MonoPair} with margins on $\mathrm{AP}|_{R_{40}}$ of +3.73\%, +2.37\%, and +1.88\%. Our method achieves first or second place on the Cyclist class with margins on $\mathrm{AP}|_{R_{40}}$ of -1.37\%, -1.33\%, and +0.18\% relative to MonoPSR~\cite{monopsr}.
\subsection{Ablation Studies} \label{Add Ablation}
\begin{table}
\small
\centering
\begin{tabular}{c|c|ccc}
\toprule
\multirow{2}{*}{Exp.} & \multirow{2}{*}{Disc. Method} & \multicolumn{3}{c}{Car (IOU = 0.7)} \\
 &  & Easy & Mod. & Hard \\ \hline
\newtag{1}{itm:dc1} & UD & 20.40 & 15.10 & 12.75 \\
\newtag{2}{itm:dc2} & SID & 22.00 & 15.65 & 13.26 \\
\newtag{3}{itm:dc3} & LID & \textbf{23.57} & \textbf{16.31} & \textbf{13.84} \\
\bottomrule
\end{tabular}
\caption{\method~Depth Discretization Ablation Experiments on the KITTI \textit{val} set using $\left.\mathrm{AP}\right|_{R_{40}}$. See Section~\ref{sec:Depth Discretization} for description of depth discretization methods.}
\label{tab:depth-disc}
\end{table}
\begin{table}
\small
\centering
\begin{tabular}{c|cc|ccc}
\toprule
\multirow{2}{*}{Exp.} & \multirow{2}{*}{Block} & \multirow{2}{*}{$W_F \times H_F$} & \multicolumn{3}{c}{Car (IOU = 0.7)} \\
 &  & &  Easy & Mod. & Hard \\ \hline
\rule{0pt}{3ex}
\newtag{1}{itm:fr1} & \textit{Block1} & $\frac{W_I}{4} \times \frac{H_I}{4}$ & \textbf{23.57} & \textbf{16.31} & \textbf{13.84} \\[1mm]
\newtag{2}{itm:fr2} & \textit{Block2} & $\frac{W_I}{8} \times \frac{H_I}{8}$ & 20.57 & 14.86 & 12.93 \\[1mm]
\newtag{3}{itm:fr3} & \textit{Block3} & $\frac{W_I}{8} \times \frac{H_I}{8}$ & 19.60 & 14.65 & 12.73 \\[1mm]
\newtag{4}{itm:fr4} & \textit{Block4} & $\frac{W_I}{8} \times \frac{H_I}{8}$ & 20.71 & 14.94 & 12.91 \\[1mm]
\bottomrule
\end{tabular}
\caption{\method~Feature Resolution Ablation Experiments on the KITTI \textit{val} set using $\left.\mathrm{AP}\right|_{R_{40}}$. Block indicates the Image Backbone block where image features $\mathbf{\tilde{F}}$ are extracted from (See Figure~\ref{fig:architecture}). {$W_F \times H_F$} indicates the width and height of image features $\mathbf{\tilde{F}}$.}
\label{tab:feat-res}
\end{table}
\noindent
\textbf{Depth Discretization}.
Table~\ref{tab:depth-disc} shows the detection peformance of~\method~with each of the depth discretization methods outlined in Section~\ref{sec:Depth Discretization}. We observe that LID offers the highest performance, leading to its use for our method.
\\[0.2\baselineskip]
\noindent
\textbf{Feature Resolution}.
Table~\ref{tab:feat-res} shows the detection peformance of~\method~when modifying the image feature extraction layer in the Image Backbone (See Figure~\ref{fig:architecture}). Experiment~\ref{itm:fr1} shows the performance when image features are extracted from \textit{Block1}. Experiments~\ref{itm:fr2},~\ref{itm:fr3}, and~\ref{itm:fr4} shows reduced performance when smaller resolution image features are extracted. Smaller spatial resolutions in the image features cause oversampling in the frustum to voxel grid transformation, leading to many voxel features $\mathbf{V}$ with similar features (See Section~\ref{sec:3D Representation Learning}).
\section{Additional Details} \label{sec:AddDetails}
\noindent
\textbf{Depth Distributions}.
We add an depth bin to our depth distributions $\mathbf{D}$ that represents any depth outside the range $[d_{min}, d_{max}]$. The added bin is included in the depth distribution loss $L_\mathrm{depth}$, but is removed when generating frustum features $\mathbf{G}$.
\\[0.2\baselineskip]
\noindent
\textbf{Training Details}.
We initialize the Image Backbone and Depth Distribution Network (See Figure~\ref{fig:architecture}) using the DeepLabV3~\cite{DeepLabV3} model pretrained on the MS-COCO dataset~\cite{COCO}. All other components are randomly initialized.
\end{document}